\documentclass{article}
\usepackage{float}

\PassOptionsToPackage{numbers, compress}{natbib}
\usepackage{tcolorbox}
\usepackage[table]{xcolor}
\usepackage{multirow}
\renewcommand{\arraystretch}{1.05}

\usepackage{amssymb}
\usepackage{amsmath}
\usepackage{newunicodechar}
\newunicodechar{✗}{\ding{55}}
\usepackage{fontawesome5}
\usepackage{pifont}
\usepackage{newunicodechar}
\newunicodechar{✓}{\checkmark}

\definecolor{helpgreen}{RGB}{60,179,113}
\definecolor{harmlessred}{RGB}{220,20,60}
\definecolor{honestblue}{RGB}{70,130,180}

\DeclareUnicodeCharacter{211D}{$\mathbb{R}$} % ℝ
\DeclareUnicodeCharacter{2264}{$\leq$}       % ≤
\DeclareUnicodeCharacter{2227}{$\wedge$}     % ∧
\DeclareUnicodeCharacter{03A3}{$\Sigma$}     % Σ
\DeclareUnicodeCharacter{222B}{$\int$}       % ∫
\DeclareUnicodeCharacter{2080}{$_0$}         % ₀
\DeclareUnicodeCharacter{0393}{$\Gamma$}     % Γ

\DeclareUnicodeCharacter{211D}{$\mathbb{R}$} % ℝ

\DeclareUnicodeCharacter{03BC}{$\mu$}     % μ
\DeclareUnicodeCharacter{03C3}{$\sigma$}  % σ
\DeclareUnicodeCharacter{03C1}{$\rho$}    % ρ
\DeclareUnicodeCharacter{03C0}{$\pi$}     % π
\DeclareUnicodeCharacter{0394}{$\Delta$}  % Δ
\DeclareUnicodeCharacter{0393}{$\Gamma$}  % Γ

\DeclareUnicodeCharacter{221A}{$\sqrt{}$} % √
\DeclareUnicodeCharacter{2248}{$\approx$} % ≈
\DeclareUnicodeCharacter{2264}{$\leq$}    % ≤
\DeclareUnicodeCharacter{2265}{$\geq$}    % ≥
\DeclareUnicodeCharacter{2227}{$\wedge$}  % ∧
\DeclareUnicodeCharacter{222B}{$\int$}    % ∫

\DeclareUnicodeCharacter{2080}{$_0$} % ₀
\DeclareUnicodeCharacter{2081}{$_1$} % ₁
\DeclareUnicodeCharacter{2082}{$_2$} % ₂
\DeclareUnicodeCharacter{2083}{$_3$} % ₃

\DeclareUnicodeCharacter{0127}{$\hbar$} % ħ

\DeclareUnicodeCharacter{0304}{\bar} % ̄

\usepackage[preprint]{neurips_2026}

\usepackage[utf8]{inputenc} % allow utf-8 input
\usepackage[T1]{fontenc}    % use 8-bit T1 fonts
\usepackage{hyperref}       % hyperlinks
\usepackage{url}            % simple URL typesetting
\usepackage{booktabs}       % professional-quality tables
\usepackage{amsfonts}       % blackboard math symbols
\usepackage{nicefrac}       % compact symbols for 1/2, etc.
\usepackage{microtype}      % microtypography
\usepackage{xcolor}         % colors

\title{We Think, Therefore We Align LLMs to Helpful, Harmless and Honest Before They Go Wrong}

\author{%
  Gautam Siddharth Kashyap \\
  School of Computing\\
  Macquarie University, Australia\\
  \texttt{gautam.kashyap@hdr.mq.edu.au} \\
  \And
  Mark Dras \\
  School of Computing\\
  Macquarie University, Australia\\
  \texttt{mark.dras@mq.edu.au} \\
  \AND
  Usman Naseem \\
  School of Computing\\
  Macquarie University, Australia\\
  \texttt{usman.naseem@mq.edu.au} \\
}

\begin{document}

\maketitle

\begin{abstract}
Alignment of Large Language Models (LLMs) is the ability to satisfy desired objectives during generation, which is critical for trustworthy deployment. In practice, alignment is often operationalized through multiple objectives such as \textit{Helpfulness}, \textit{Harmlessness}, and \textit{Honesty} (HHH). Prior works study alignment \textit{via} steering vectors in standard Transformer decoders but treat objectives in isolation, where optimizing a single objective can overwrite others, leading to interference. Recent works attempt to address this limitation by extending steering to a 1-to-N Transformer setting (i.e., a shared representation replicated into multiple objective-specific pathways within a single forward pass) by replicating representations into objective-specific pathways, but apply transformations independently, resulting in inconsistent responses across objectives. Similarly, approaches such as safe RLHF and MoE-based designs study trade-offs across objectives but do not constrain objective-specific transformations within a shared representation during inference. As a result, even aligned State-of-the-Art (SOTA) LLMs can struggle to jointly satisfy HHH objectives in complex settings. To address this, we propose \textbf{\textit{Adaptive Multi-Branch Steering (AMBS)}}, a two-stage framework in a 1-to-N Transformer setting that parameterizes objective-specific transformations relative to a shared representation. In Stage I, a shared hidden representation is computed once. In Stage II, this representation is replicated into $N$ pathways and updated relative to a shared reference, capturing objective-specific deviations while restricting divergence. This produces $N$ objective-specific responses within a single forward pass, which can be combined at decoding to obtain a single response across objectives. Across multiple backbones, AMBS improves performance across HHH, with consistent gains in WR, TI, and SS (e.g., Avg $56.5\%$ on \texttt{LLaMA-2-7B}) while maintaining efficiency (e.g., $189$ Tok/s, $9$ GPU-hrs).
\end{abstract}

\begin{figure*}[t!]
\centering
\scriptsize

% Shared prompt
\begin{tcolorbox}[colback=gray!5, colframe=blue!60!black, width=0.95\textwidth,
    boxsep=2pt, left=2pt, right=2pt, top=2pt, bottom=2pt, title=Shared Prompt (User)]
\textbf{\faUser\ Instruction:} \textit{Explain why climate change is important.}
\end{tcolorbox}

\vspace{-0.2cm}

\begin{minipage}[t][0.55\textheight][t]{0.45\textwidth} % fixed height
\vspace{0pt} % ensures top alignment
\begin{tcolorbox}[colback=gray!3, colframe=violet!80!black, width=\textwidth,
    boxsep=2pt, left=2pt, right=2pt, top=2pt, bottom=2pt, title=Naïve Steering via 1-to-N Transformer (Incorrect)]
{\textcolor{helpgreen}{\faRobot Helpfulness:}} \textit{"Climate change is bad. Deal with it."} {\textcolor{helpgreen}{✗}}\\
{\textcolor{harmlessred}{\faRobot Harmlessness:}} \textit{"Burn waste to reduce climate change."} {\textcolor{harmlessred}{✗}}\\
{\textcolor{honestblue}{\faRobot Honesty:}} \textit{"Climate change is a myth created by scientists."} {\textcolor{honestblue}{✗}}
\end{tcolorbox}

\begin{tcolorbox}[colback=gray!3, colframe=purple!80!black, width=\textwidth,
    boxsep=2pt, left=2pt, right=2pt, top=2pt, bottom=2pt, title=AMBS via 1-to-N Transformer (Correct)]
{\textcolor{helpgreen}{\faRobot Helpfulness:}} \textit{"Climate change harms health."} {\textcolor{helpgreen}{✓}}\\
{\textcolor{harmlessred}{\faRobot Harmlessness:}} \textit{"Cut fossil fuels and plant trees safely."} {\textcolor{harmlessred}{✓}}\\
{\textcolor{honestblue}{\faRobot Honesty:}} \textit{"NASA confirms warming is due to greenhouse gases."} {\textcolor{honestblue}{✓}}
\end{tcolorbox}
\vfill % pushes content down to align with right side
\end{minipage}
\hfill
\begin{minipage}[t][0.55\textheight][t]{0.5\textwidth}
\vspace{0pt}
\centering
\includegraphics[width=\textwidth]{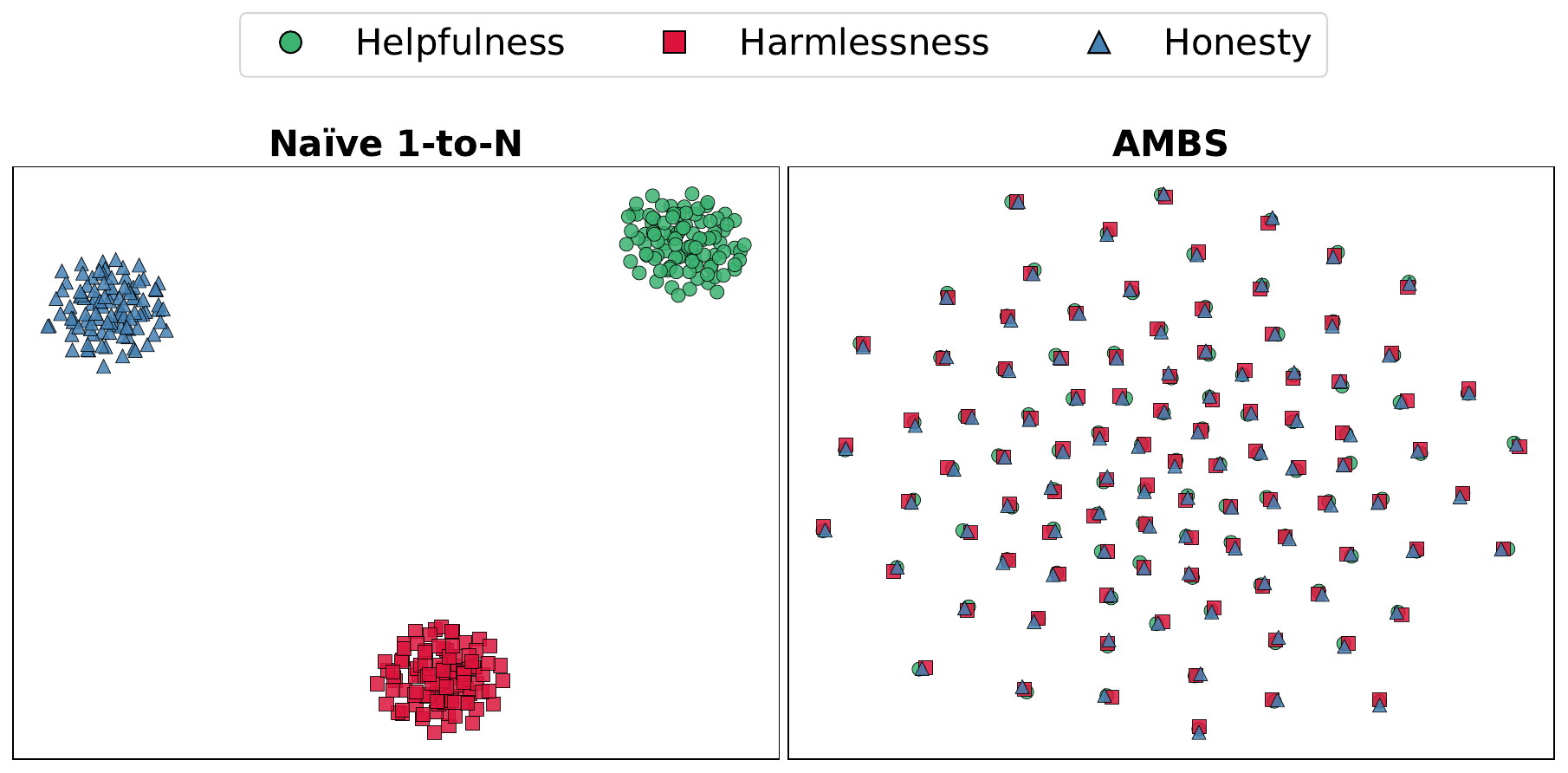}
\vfill
\end{minipage}

\vspace{-8.7cm}

\caption{Motivation for AMBS in multi-objective alignment. \textbf{Top}: A shared prompt. \textbf{Left}: In a 1-to-N Transformer, $h_\ell$ is replicated into pathways $\{h_\ell^{(i)}\}_{i=1}^N$. Naïve steering applies independent updates $\tilde{h}_\ell^{(i)} = h_\ell + \Delta h_\ell^{(i)}$, leading to divergence across objectives. AMBS uses a shared reference $h_\ell^{\text{ref}} = h_\ell^{(0)}$ with $\tilde{h}_\ell^{(i)} = h_\ell^{\text{ref}} + \Delta h_\ell^{(i)}$, where $\Delta h_\ell^{(i)} = f_i(\tilde{h}_\ell^{(i)} - h_\ell^{\text{ref}})$, so pathways differ only through deviation terms and are combined at decoding into a single response. \textbf{Right}: t-SNE of $\tilde{h}_\ell^{(i)}$ at $\ell = L$ (\texttt{LLaMA-2-7B}), aggregated across samples (perplexity = 25). Naïve pathways form separated clusters, AMBS shows greater overlap, indicating reduced divergence.}
\label{fig:motivation_hhh_with_tsne}
\vspace{-0.6cm}
\end{figure*}

\section{Introduction}
\label{Introduction}

Alignment of Large Language Models (LLMs) refers to the ability to satisfy desired objectives during generation, which is essential for trustworthy deployment \cite{naseem2025alignment}. Formally, given an input $x$ and generated response $y \sim p_\theta(y \mid x)$, alignment requires that $y$ satisfies a set of objectives $\mathcal{O} = \{o_1, o_2, \dots, o_k\}$ defined by human intent \cite{ji2026survey}, typically encoded through Supervised Fine-Tuning (SFT) \cite{li2024getting} or Reinforcement Learning from Human Feedback (RLHF) \cite{barnhart2025aligning}. In practice, alignment is studied as a multi-objective problem involving \textit{Helpfulness}, \textit{Harmlessness}, and \textit{Honesty} (HHH) \cite{askell2021general}, where a \textit{Helpful} response provides relevant and actionable information, a \textit{Harmless} response avoids unsafe or toxic information, and an \textit{Honest} response maintains factual information. These objectives are interdependent and may conflict \cite{bai2022training}, e.g., a \textit{Helpful} response to a sensitive input must also remain \textit{Harmless}, leading to the goal of satisfying $\mathcal{O}_{\text{HHH}} = \{o_{\text{help}}, o_{\text{harm}}, o_{\text{hon}}\}$ during generation with consistency across objectives rather than optimizing each independently. 

To achieve alignment, prior works employ steering vectors (e.g., \cite{turner2023steering, elhage2022toy, subramani2022extracting}) injected into hidden states of standard Transformer decoders, where a single forward pass produces a single response distribution. Given a hidden representation $h_\ell$ at layer $\ell$, a steering vector $\Delta h_\ell$ is added such that $\tilde{h}_\ell = h_\ell + \Delta h_\ell$, mapping the model toward a target objective. While effective for single-objective control, this treats objectives in isolation, where optimizing one objective alters shared representations and can overwrite others, leading to interference within the same pathway. Recent works (e.g., \cite{nguyen2025multi, tan2024analysing}) extend steering to a 1-to-N Transformer setting, where a shared representation is replicated into multiple objective-specific pathways within a single forward pass; however, independent transformations reduce direct interference but do not constrain relationships across pathways, resulting in inconsistent responses across objectives (see Figure~\ref{fig:motivation_hhh_with_tsne}). In parallel, approaches such as safe RLHF \cite{tekin2025dynamic} and MoE-based designs \cite{kashyap2026model, tekin2026h3fusion, kashyap2025too} study trade-offs across objectives through training-time optimization or modular architectures, but do not constrain objective-specific transformations within a shared representation during inference. As a result, even aligned State-of-the-Art (SOTA) LLMs (e.g., \cite{bai2023qwen, touvron2023llama}) can struggle to jointly satisfy HHH objectives in complex settings, highlighting a gap in multi-objective alignment at inference time, where one seeks transformations $\{\Delta h_\ell^{(i)}\}_{i=1}^N$ that satisfy each objective while maintaining agreement across pathways.

To operationalize this formulation, we propose \textbf{\textit{Adaptive Multi-Branch Steering (AMBS)}}, a two-stage framework for multi-objective alignment in a 1-to-N Transformer setting\footnote{\scriptsize{We use \emph{representation} term to refer to the hidden states $h_\ell \in \mathbb{R}^{T \times d}$ at layer $\ell$. A \emph{pathway} term refers a replicated copy for objective $o_i$, i.e., $h_\ell^{(i)}$. A \emph{transformation} term refers to the objective-specific update $\Delta h_\ell^{(i)}$ applied to a pathway.}}. In Stage I, the Transformer computes hidden representations $\{h_\ell\}_{\ell=1}^L$ and defines a shared representation $h_\ell^{(0)} = h_\ell$, reused across objectives. This representation is replicated into $N$ pathways as $h_\ell^{(i)} = h_\ell^{(0)}, \ i = 1, \dots, N$. In Stage II, each pathway applies an objective-specific transformation $\Delta h_\ell^{(i)}$ relative to the shared reference $h_\ell^{\text{ref}} = h_\ell^{(0)}$, producing $\tilde{h}_\ell^{(i)} = h_\ell^{\text{ref}} + \Delta h_\ell^{(i)}$. The transformed representations are mapped to output distributions $\{p_\theta^{(i)}\}_{i=1}^N$ within a single forward pass and combined at decoding to produce a single response across objectives. Divergence across pathways is controlled by
$\min_{\{\Delta h_\ell^{(i)}\}} \sum_{i=1}^N \mathcal{L}_{\text{obj}}^{(i)} + \lambda \mathcal{L}_{\text{cons}}$, where $\mathcal{L}_{\text{obj}}^{(i)}$ is the objective-specific loss and $\mathcal{L}_{\text{cons}}$ measures divergence. In summary, the main contribution of this work are as follows:

\begin{itemize}
\vspace{-0.3cm}
   \item We study multi-objective alignment in LLMs \textit{via} \textbf{\textit{AMBS}}, a two-stage framework that extends a shared representation to a 1-to-N Transformer setting with objective-specific transformations $\{\Delta h_\ell^{(i)}\}_{i=1}^N$, formulated as learning these transformations under a shared decoding scheme while controlling loss and divergence across pathways for $\mathcal{O}_{\text{HHH}}$. 
    \vspace{-0.18cm}
   \item Empirically, AMBS improves joint HHH performance across four backbones, achieving higher Avg (e.g., $56.5\%$ vs.\ $55.1\%$ on \texttt{LLaMA-2-7B}) while reducing unsafe outputs and maintaining comparable throughput and compute cost.
\end{itemize}

\section{Related Works}
\label{RelatedWork}

\paragraph{General Alignment in LLMs.}

Alignment in LLMs has been primarily studied through training-based approaches as discussed in \S\ref{Introduction}, where models such as \texttt{InstructGPT} \cite{ouyang2022training}, \texttt{ChatGPT} \cite{achiam2023gpt}, \texttt{LLaMA-2-Chat} \cite{touvron2023llama}, \texttt{Mistral-Instruct} \cite{jiang2023mistral7b}, and \texttt{Qwen-Instruct} \cite{bai2023qwen} apply SFT followed by RLHF to align responses with human intent, while alternatives such as RAHF \cite{liu2024aligning} and Aligner \cite{ji2024aligner} study representation-level alignment. These approaches optimize a scalar reward $r(y, x)$ during training, but require costly labeling and provide limited control at inference time. Recent work studies inference-time alignment via steering vectors and representation editing, where hidden states are modified as $\tilde{h}_\ell = h_\ell + \Delta h_\ell$, including PPLM \cite{dathathri2019plug}, DExperts \cite{liu2021dexperts}, activation steering \cite{turner2023steering}, ReFT \cite{wu2024reft}, and LoRA \cite{hu2022lora}-based editing. Extensions compose multiple directions in representation space, such as SRS \cite{he2025towards} and persona or code-style vectors \cite{konen2024style, bayat2025steering}. Despite these advances, prior approaches operate in a single-objective setting, where a single transformation $\Delta h_\ell$ is applied per forward pass or multiple directions are composed without constraints, leading to interference within a shared pathway and unstable behavior when combining objectives (see \S\ref{Introduction}).

\vspace{-0.4cm}

\paragraph{HHH Alignment in LLMs}

Alignment along HHH has been studied using datasets and benchmarks such as \texttt{Alpaca} \cite{taori2023stanford}, \texttt{BeaverTails} \cite{ji2023beavertails}, and \texttt{TruthfulQA} \cite{lin2022truthfulqa}. Prior works on multi-objective alignment include TrinityX \cite{kashyap2025too}, which introduces a Mixture-of-Calibrated-Experts (MoCaE) framework with separate experts for each HHH objective and calibrated routing, and AlignX \cite{kashyap2026model}, which identifies failure modes such as \textit{axis collapse} where conflicting objectives lead to disjoint representations and unreliable routing, and proposes a two-stage framework to separate objective-specific features and improve routing stability. H3Fusion \cite{tekin2026h3fusion} formulates alignment as a fusion problem in representation space using a drift-based objective, MARL-Focal \cite{tekin2025dynamic} adopts a multi-agent setting where multiple LLM agents collaborate and resolve objectives through reward-aware fusion, and STARS \cite{zhu2026exploring} applies multi-branch activation steering to generate diverse responses across directions in hidden space. However, these approaches rely on independently optimized objectives and do not constrain how objective-specific transformations relate within a shared representation during inference, leading to inconsistencies across objectives. In contrast, we study coordinated steering in a 1-to-N Transformer setting, where a shared representation $h_\ell^{(0)}$ is replicated across $N$ pathways and objective-specific transformations $\{\Delta h_\ell^{(i)}\}_{i=1}^N$ are defined relative to a common reference $h_\ell^{\text{ref}} = h_\ell^{(0)}$. 

\begin{figure*}[t!]
    \centering
    \includegraphics[width=0.9\textwidth]{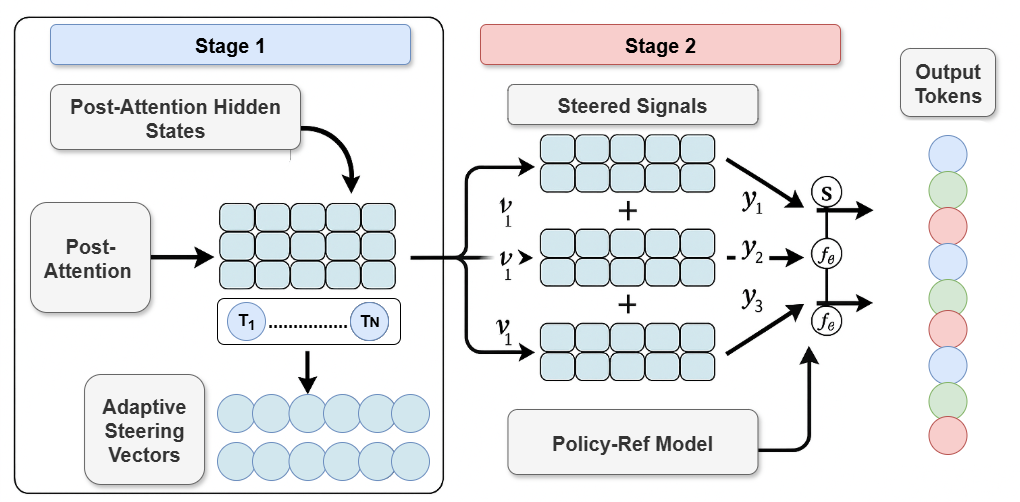} 
    \vspace{-0.3cm}
    \caption{Overview of \textit{\textbf{Adaptive Multi-Branch Steering (AMBS)}} in a 1-to-N Transformer setting. Stage I computes shared hidden representations $h_\ell^{(0)}$ as a common basis. Stage II replicates this representation into $N$ pathways $h_\ell^{(i)}$, applies reference-relative transformations $\tilde{h}_\ell^{(i)} = h_\ell^{\text{ref}} + \Delta h_\ell^{(i)}$, and produces objective-specific responses $\{p_\theta^{(i)}\}_{i=1}^N$ in a single forward pass, which are combined at decoding into a single response across objectives.}
    \label{figure2}
    \vspace{-0.3cm}
\end{figure*}

\section{Methodology}
\label{Methodology}

Unlike prior approaches that parameterize $\Delta h_\ell^{(i)}$ independently as functions of $h_\ell$, AMBS defines transformations as functions of deviations $(h_\ell^{(i)} - h_\ell^{\text{ref}})$ relative to a shared reference in a two-stage framework (see Figure~\ref{figure2}): (i) compute a shared representation $h_\ell^{(0)}$ as a common basis (Stage I), and (ii) apply reference-guided transformations $\{\Delta h_\ell^{(i)}\}$ relative to this state (Stage II), producing $N$ objective-specific responses $\{p_\theta^{(i)}\}_{i=1}^N$ within a single forward pass.

\subsection{Shared Representation}
\label{Shared Representation}

Given an input prompt $x = (x_1, \dots, x_T)$ from an HHH-aligned dataset (see \S\ref{Dataset}), where $x_t \in \mathcal{V}$ for $t = 1, \dots, T$, $\mathcal{V}$ is the vocabulary and $T$ is the sequence length, we map tokens to embeddings as $h_0 = E(x) + P$, where $E \in \mathbb{R}^{|\mathcal{V}| \times d}$ is the token embeddings matrix, $P \in \mathbb{R}^{T \times d}$ is the positional embeddings matrix, $d$ is the hidden dimension, and $h_0 \in \mathbb{R}^{T \times d}$ is the input representation. A Transformer decoder processes these representation through $L$ layers as $h_\ell = \mathrm{Transformer}_\ell(h_{\ell-1}), \ \ell = 1, \dots, L$, where $h_\ell \in \mathbb{R}^{T \times d}$ is the hidden representation at layer $\ell$, with Masked Self-Attention ($\mathrm{MSA}_\ell$) and Feed-Forward Network ($\mathrm{FFN}_\ell$) with residual connections as per Equation (1).
\[
\hat{h}_\ell = \mathrm{MSA}_\ell(h_{\ell-1}) + h_{\ell-1}, \quad
h_\ell = \mathrm{FFN}_\ell(\hat{h}_\ell) + \hat{h}_\ell \tag{1}
\]
We use $h_\ell$ (in practice $\ell = L$) as the basis for steering (see \S\ref{Ablation Study}) and define the shared representation $h_\ell^{(0)} = h_\ell$, computed once per forward pass and reused across objectives. In a 1-to-N Transformer setting, it is replicated into $N$ pathways as per Equation (2).
\[
h_\ell^{(i)} = h_\ell^{(0)}, \quad i = 1, \dots, N \tag{2}
\]

\subsection{Shared Steering}
\label{Shared Steering}

Given the replicated representations $\{h_\ell^{(i)}\}_{i=1}^N$ from Equation (2), where $h_\ell^{(i)} \in \mathbb{R}^{T \times d}$ denotes the hidden representation for pathway $i$ at layer $\ell$, we apply objective-specific transformations toward each objective $o_i \in \mathcal{O}_{\text{HHH}}$. We define a shared reference $h_\ell^{\text{ref}} = h_\ell^{(0)}$, where $h_\ell^{(0)}$ is the representation computed in Stage I and reused across pathways. For each objective $o_i$, a policy $f_i(\cdot; \phi_i)$ specifies how the representation is updated relative to this reference. Each pathway applies a transformation conditioned on the deviation from the shared reference as per Equations (3) and (4)\footnote{\scriptsize{Although Equations (3) and (4) are written in a self-referential form, the update is computed in a single forward pass with an order. We first construct an intermediate state $\tilde{h}_\ell^{(i,0)} = h_\ell^{\text{ref}} + b_i$, yielding a non-zero deviation $(\tilde{h}_\ell^{(i,0)} - h_\ell^{\text{ref}}) = b_i$. This is used to compute $\Delta h_\ell^{(i)} = f_i(\tilde{h}_\ell^{(i,0)} - h_\ell^{\text{ref}}; \phi_i)$, and the final representation is $\tilde{h}_\ell^{(i)} = h_\ell^{\text{ref}} + \Delta h_\ell^{(i)}$. No fixed-point computation or iteration is used.}}, defined in terms of $(\tilde{h}_\ell^{(i)} - h_\ell^{\text{ref}})$ rather than independent updates.
\[
\Delta h_\ell^{(i)} = f_i\big(\tilde{h}_\ell^{(i)} - h_\ell^{\text{ref}}; \phi_i\big), \quad \Delta h_\ell^{(i)} \in \mathbb{R}^{T \times d} \tag{3}
\]
\[
\tilde{h}_\ell^{(i)} = h_\ell^{\text{ref}} + \Delta h_\ell^{(i)}, \quad \tilde{h}_\ell^{(i)} \in \mathbb{R}^{T \times d} \tag{4}
\]
In practice, the policy $f_i(\cdot)$ is implemented as a low-rank projection with an additive bias applied to the deviation $(\tilde{h}_\ell^{(i,0)} - h_\ell^{\text{ref}})$ as per Equation (5). The transformation is applied along the hidden dimension $d$ for each token position, and at the final layer ($\ell = L$), which controls the output distribution (see \S\ref{Ablation Study}). Although the base model is frozen, the initial bias produces a non-zero deviation $(\tilde{h}_\ell^{(i,0)} - h_\ell^{\text{ref}}) = b_i$, which is used to compute the low-rank update in the same forward pass (see Lemma 2). Specifically, $A_i \in \mathbb{R}^{d \times r}$ and $B_i \in \mathbb{R}^{r \times d}$ are learnable low-rank matrices, and $b_i \in \mathbb{R}^{d}$ is a learnable bias (see Lemma 1), where $r \ll d$ controls the capacity of the transformation. Smaller $r$ restricts updates to a lower-dimensional subspace, while larger $r$ increases flexibility (see \S\ref{Ablation Study}). This parameterization captures objective-specific variations under reference-relative updates\footnote{\scriptsize{The parameters $\{A_i, B_i, b_i\}_{i=1}^N$ are randomly initialized and learned by minimizing Equation (10) with the base model fixed. The computation starts from $\tilde{h}_\ell^{(i,0)} = h_\ell^{\text{ref}} + b_i$, producing a non-zero deviation that activates the low-rank components within a single forward pass (see Lemmas 1--2).}}
\[
\Delta h_\ell^{(i)} = \big((\tilde{h}_\ell^{(i,0)} - h_\ell^{\text{ref}}) A_i\big) B_i + b_i \tag{5}
\]
From Equations (3) and (4), since $h_\ell^{\text{ref}}$ is shared across pathways, each $\tilde{h}_\ell^{(i)}$ is expressed as a common reference plus an objective-specific deviation. Differences across pathways therefore depend only on the deviation terms as per Equation (6), rather than independent transformations of $h_\ell$. The transformed representation $\tilde{h}_\ell^{(i)} \in \mathbb{R}^{T \times d}$ is mapped to the output distribution through the shared language modeling head as per Equation (7), where $\tilde{h}_\ell^{(i,t)} \in \mathbb{R}^{d}$ denotes the hidden state at position $t$ for pathway $i$, and $W_o \in \mathbb{R}^{|\mathcal{V}| \times d}$ is the shared output projection matrix.
\[
\tilde{h}_\ell^{(i)} - \tilde{h}_\ell^{(j)} = \Delta h_\ell^{(i)} - \Delta h_\ell^{(j)} \tag{6}
\]
\[
p_\theta^{(i)}(y_t \mid x, y_{<t}) = \mathrm{softmax}(W_o \tilde{h}_\ell^{(i,t)}) \tag{7}
\]
At inference, the model produces $N$ objective-specific output distributions $\{p_\theta^{(i)}\}_{i=1}^N$ in parallel within a single forward pass, where each pathway maintains its own transformed representation $\tilde{h}_\ell^{(i)}$ from Equation (4) and is mapped to the output distribution as per Equation (7), i.e., $p_\theta^{(i)}(y_t \mid x, y_{<t}) = \mathrm{softmax}(W_o \tilde{h}_\ell^{(i,t)})$ for $i = 1, \dots, N$. The additional computation scales linearly with $N$ through parallel pathway projections while sharing the base Transformer computation (see \S\ref{Ablation Study}). The outputs $\{p_\theta^{(i)}\}_{i=1}^N$ can be retained as objective-specific responses or combined into a single response across objectives. We aggregate the distributions into a shared decoding distribution: $p_\theta^{\text{joint}}(y_t \mid x, y_{<t}) \propto \exp\left( \sum_{i=1}^N w_i \log p_\theta^{(i)}(y_t \mid x, y_{<t}) \right)$, and perform decoding using $p_\theta^{\text{joint}}$ (see \S\ref{sec:joint_empirical}). For agreement across pathways, we include a consistency term over the transformed representations at the token level as per Equation (8) (see \S\ref{Ablation Study}), which penalizes divergence between pathway representations in the deviation space relative to $h_\ell^{\text{ref}}$, without enforcing equality due to competing objective-specific gradients (see Lemma 3).
\[
\mathcal{L}_{\text{cons}} = \frac{1}{T} \sum_{t=1}^T \sum_{i < j} \left\| \tilde{h}_\ell^{(i,t)} - \tilde{h}_\ell^{(j,t)} \right\|_2^2\tag{8}
\]
The overall objective is given in Equation (10), where $\mathcal{L}_{\text{obj}}^{(i)}$ denotes the intra-objective loss for objective $o_i$ applied to $p_\theta^{(i)}$, and $\mathcal{L}_{\text{cons}}$ captures divergence across pathways at the representation level (see Lemma 3). Each objective-specific loss is defined over the corresponding output distribution using token-level supervision from objective-specific labels in the HHH-aligned dataset (see \S\ref{Ablation Study})\footnote{\scriptsize{The supervision signals $\{y^{(i)}\}_{i=1}^N$ correspond to objective-specific labels from the HHH-aligned dataset. $\lambda$ is selected via validation (see \S\ref{Ablation Study}) and controls agreement across pathways.}}. For datasets with classification-style labels (e.g., BeaverTails), labels $z \in \{0,1\}$ are converted into token-level supervision by mapping each instance to a short target sequence $y^{(i)} = g(z)$ (e.g., safe refusal for $z=0$, compliant answer for $z=1$). The loss in Equation (9) is then applied over the tokens of $y^{(i)}$, yielding sequence-level supervision under the language modeling objective.
\[
\mathcal{L}_{\text{obj}}^{(i)} = - \sum_{t=1}^T \log p_\theta^{(i)}(y_t^{(i)} \mid x, y_{<t}^{(i)})\tag{9}
\]
\vspace{-0.5cm}
\[
\min_{\{\Delta h_\ell^{(i)}\}} \sum_{i=1}^N \mathcal{L}_{\text{obj}}^{(i)} + \lambda \mathcal{L}_{\text{cons}} \tag{10}
\]

\begin{figure}[h]
\vspace{-0.6cm}
\tiny
\centering
\begin{tcolorbox}[
colback=blue!2,
colframe=black!30,
boxsep=2pt,
left=3pt,right=3pt,top=3pt,bottom=3pt,
arc=2pt,
width=0.95\linewidth
]

\textbf{Lemma 1 (Non-Degenerate Initialization).}
Let the base Transformer be frozen. With $\tilde{h}_\ell^{(i)} = h_\ell^{\text{ref}}$ at initialization and $\Delta h_\ell^{(i)} = ((\tilde{h}_\ell^{(i)} - h_\ell^{\text{ref}}) A_i) B_i + b_i$, we have $\Delta h_\ell^{(i)} = b_i \neq 0$ almost surely, hence $\tilde{h}_\ell^{(i)} \neq h_\ell^{\text{ref}}$ (see \S\ref{Ablation Study}). \\
\vspace{-2pt}
\textit{Proof.}
At initialization, $\tilde{h}_\ell^{(i)} - h_\ell^{\text{ref}} = 0$. Substituting into Equation (5) gives $\Delta h_\ell^{(i)} = b_i$. Under random initialization $b_i \sim \mathcal{D}$, we have $b_i \neq 0$ almost surely, hence $\tilde{h}_\ell^{(i)} = h_\ell^{\text{ref}} + b_i \neq h_\ell^{\text{ref}}$. \hfill $\square$

\vspace{2pt}

\textbf{Lemma 2 (Immediate Activation of Low-Rank Parameters).}
Assume the base model is frozen and optimization is performed over $\{A_i, B_i, b_i\}$ via Equation (10), where $\Delta h_\ell^{(i)} = f_i(\tilde{h}_\ell^{(i)} - h_\ell^{\text{ref}})$. Then at initialization, $\nabla_{b_i} \neq 0$, and within the same forward computation, $\nabla_{A_i}, \nabla_{B_i} \neq 0$.

\vspace{-2pt}
\textit{Proof.}
The loss depends on parameters through $\tilde{h}_\ell^{(i)}$, with gradient flow $\mathcal{L} \rightarrow \tilde{h}_\ell^{(i)} \rightarrow \Delta h_\ell^{(i)} \rightarrow \{A_i, B_i, b_i\}$. The computation starts from $\tilde{h}_\ell^{(i)} = h_\ell^{\text{ref}} + b_i$, giving $\frac{\partial \tilde{h}_\ell^{(i)}}{\partial b_i} = I$, hence $\nabla_{b_i} \neq 0$. This produces a non-zero deviation $\tilde{h}_\ell^{(i)} - h_\ell^{\text{ref}} = b_i \neq 0$, so $\Delta h_\ell^{(i)}$ depends on $A_i$ and $B_i$, implying $\nabla_{A_i}, \nabla_{B_i} \neq 0$. The role of $b_i$ induces a non-zero deviation $(\tilde{h}_\ell^{(i)} - h_\ell^{\text{ref}} = b_i)$, activating the low-rank update and for gradients to $A_i,B_i$ in one forward pass; at inference, it remains a fixed offset keeping the transformation active.
\hfill $\square$

\vspace{2pt}

\textbf{Lemma 3 (Consistency--Objective Trade-off at Stationarity).}
Let $\tilde{h}_\ell^{(i)}$ be a stationary point of Equation (10) with respect to $\tilde{h}_\ell^{(i,t)}$. Then for each $i,t$, 
$\nabla_{\tilde{h}_\ell^{(i,t)}} \mathcal{L}_{\text{obj}}^{(i)} 
+ \frac{2\lambda}{T} \sum_{j \neq i} (\tilde{h}_\ell^{(i,t)} - \tilde{h}_\ell^{(j,t)}) = 0$, 
which characterizes the balance between objective gradients and pairwise coupling terms (see \S\ref{Ablation Study}). \\
\vspace{-2pt}
\textit{Proof.}
From Equation (10), $\mathcal{L} = \sum_{i=1}^N \mathcal{L}_{\text{obj}}^{(i)} + \lambda \mathcal{L}_{\text{cons}}$, where $\mathcal{L}_{\text{cons}} = \frac{1}{T} \sum_{t=1}^T \sum_{i<j} \|\tilde{h}_\ell^{(i,t)} - \tilde{h}_\ell^{(j,t)}\|_2^2$. At a stationary point, $\nabla_{\tilde{h}_\ell^{(i,t)}} \mathcal{L} = 0$. The objective term contributes $\nabla_{\tilde{h}_\ell^{(i,t)}} \mathcal{L}_{\text{obj}}^{(i)}$, and the consistency term gives $\frac{2}{T} \sum_{j \neq i} (\tilde{h}_\ell^{(i,t)} - \tilde{h}_\ell^{(j,t)})$. Combining both terms yields the stated result.
\hfill $\square$

\end{tcolorbox}
\vspace{-0.3cm}
\end{figure}

\section{Experimental Setup}
\label{Experimental Setup}

\paragraph{Dataset.}
\label{Dataset}

To evaluate multi-objective alignment along HHH, we use standard benchmarks for each objective as in prior works \cite{kashyap2026model, tekin2026h3fusion, kashyap2025too}. For \textit{Helpfulness}, we use \texttt{Alpaca}\footnote{\scriptsize{\url{https://github.com/tatsu-lab/stanford_alpaca}}} \cite{taori2023stanford}, with 20,000 training instances and 805 held-out instances. For \textit{Harmlessness}, we use \texttt{BeaverTails}\footnote{\scriptsize{\url{https://sites.google.com/view/pku-beavertails}}} \cite{ji2023beavertails}, with 27,186 safe instances for training and 3,021 unsafe instances for testing. For \textit{Honesty}, we use \texttt{TruthfulQA}\footnote{\scriptsize{\url{https://github.com/sylinrl/TruthfulQA}}} \cite{lin2022truthfulqa}, where supervision is constructed by expanding labeled responses into training instances and testing is performed on a held-out subset with no question overlap. All datasets follow the splits provided in their respective sources. Each dataset corresponds to an objective in $\mathcal{O}_{\text{HHH}}$, and supervision is applied independently per objective (see \S\ref{Ablation Study}).

\vspace{-0.2cm}

\paragraph{Evaluation Metrics.}
\label{Evaluation Metrics}

We evaluate alignment along HHH using objective-specific metrics as in prior works \cite{kashyap2026model, tekin2026h3fusion, kashyap2025too}. For each objective $o_i \in \mathcal{O}_{\text{HHH}}$, metrics are computed over responses from the corresponding pathway distribution $p_\theta^{(i)}$ in the single-objective setting. For joint HHH evaluation, a single response is generated using $p_\theta^{\text{joint}}$, and all metrics are computed on this shared response. \textit{Helpfulness} is measured via Win Rate (WR$\uparrow$), defined as $\mathrm{WR} = \frac{\#\text{wins}}{\#\text{samples}} \times 100$, using \texttt{GPT-4o}\footnote{\scriptsize{\url{https://openrouter.ai/openai/gpt-4o}}}. \textit{Harmlessness} is measured using \texttt{Beaver-Dam-7B}\footnote{\scriptsize{\url{https://huggingface.co/PKU-Alignment/beaver-dam-7b}}}, reporting Safety Score (SS$\downarrow$) as $\mathrm{SS} = \frac{\#\text{unsafe}}{\#\text{samples}} \times 100$. \textit{Honesty} is evaluated using Truthfulness (T) and Informativeness (I) via \texttt{GPT-4o}, combined as $\mathrm{TI} = \left(\frac{\#\text{truthful}}{\#\text{samples}}\right)\left(\frac{\#\text{informative}}{\#\text{samples}}\right) \times 100$. We report a composite score $\text{Avg} = (\text{WR} + \text{TI} - \text{SS})/3$ as a coarse summary; due to differing scales and judges, conclusions are based on per-objective metrics alongside Avg, with $\uparrow$ indicating higher-is-better and $\downarrow$ indicating lower-is-better (see \S\ref{Ablation Study} for human evaluation). 

\vspace{-0.4cm}
\paragraph{Hyperparameters.}
\label{Hyperparameters}

We train the transformation parameters $\{A_i, B_i, b_i\}_{i=1}^N$ in Eq. (5), keeping the base Transformer and output projection $W_o$ (Eq. (7)) frozen, and optimize Eq. (10), with gradients propagating from Eq. (9) through Eq. (7) to $\tilde{h}_\ell^{(i)}$ (Eqs. (4)--(5)). Each dataset corresponds to an objective in $\mathcal{O}_{\text{HHH}}$, with supervision applied independently per objective (see \S\ref{Ablation Study}). We set $N=3$, $r=8$, and apply transformations at the final layer ($\ell = L$, where $L=32$ for \texttt{LLaMA-2-7B}, \texttt{Mistral-7B}, and \texttt{DeepSeek-7B}, and $L=28$ for \texttt{Gemma-7B}), with $\lambda=0.1$. Optimization uses AdamW with $\eta=2\times10^{-4}$, $\beta_1=0.9$, $\beta_2=0.999$, and weight decay $0.01$, trained for 3 epochs with batch size $B=16$ and sequence length $T=512$ in FP16 on a single NVIDIA A100 (80GB). At inference, we use equal weights $w_i = \frac{1}{N}$ for joint decoding with temperature $1.0$ and top-$p=0.9$. Hyperparameter sensitivity of $r$, $\lambda$, $L$, and optimization parameters ($\eta$, $B$) is reported in \S\ref{Hyperparameter Sensitivity Analysis}.

\vspace{-0.4cm}
\paragraph{Baselines.}
\label{Baselines}

We compare AMBS on four LLM backbones \texttt{LLaMA-2-7B}\footnote{\scriptsize{\url{https://huggingface.co/meta-llama/Llama-2-7b}}}, \texttt{Mistral-7B}\footnote{\scriptsize{\url{https://huggingface.co/mistralai/Mistral-7B-v0.1}}}, \texttt{Gemma-7B}\footnote{\scriptsize{\url{https://huggingface.co/google/gemma-7b}}}, and \texttt{DeepSeek-7B}\footnote{\scriptsize{\url{https://huggingface.co/deepseek-ai/deepseek-llm-7b-base}}}. Single-objective and inference-time baselines follow implementations as per \S\ref{Hyperparameters}, including RAHF~\cite{liu2024aligning} (\textit{Helpfulness}), Aligner~\cite{ji2024aligner} (\textit{Harmlessness}, \textit{Honesty}), and PPLM~\cite{dathathri2019plug}, DExperts~\cite{liu2021dexperts}, and activation steering~\cite{turner2023steering}. For joint HHH alignment, we compare with MARL-Focal~\cite{tekin2025dynamic}, AlignX~\cite{kashyap2026model}, TrinityX~\cite{kashyap2025too}, and H$^3$Fusion~\cite{tekin2026h3fusion}, reporting results from their original sources.

\begin{table*}[t]
\caption{Evaluation across HHH objectives with SOTAs. For each backbone, we report Win Rate (WR$\uparrow$), Safety Score (SS$\downarrow$), Truthfulness-Informativeness (TI$\uparrow$), overall score (Avg$\uparrow$), and compute metrics (Tok/s$\uparrow$ (Tk), Mem$\downarrow$ (Me), GPU-hrs$\downarrow$ (GP)).}

\label{tab:main_hhh_compute}
\centering
\tiny
\setlength{\tabcolsep}{1.5pt}
\renewcommand{\arraystretch}{0.95}

\begin{tabular}{@{}l
rrrrrrr
rrrrrrr
rrrrrrr
rrrrrrr@{}}
\toprule
\multirow{2}{*}{\textbf{Method}} 
& \multicolumn{7}{c}{\textbf{LLaMA-2-7B}} 
& \multicolumn{7}{c}{\textbf{Mistral-7B}} 
& \multicolumn{7}{c}{\textbf{Gemma-7B}} 
& \multicolumn{7}{c}{\textbf{DeepSeek-7B}} \\
\cmidrule(lr){2-8} \cmidrule(lr){9-15} \cmidrule(lr){16-22} \cmidrule(lr){23-29}
& WR & SS & TI & Avg & Tk & Me & GP
& WR & SS & TI & Avg & Tk & Me & GP
& WR & SS & TI & Avg & Tk & Me & GP
& WR & SS & TI & Avg & Tk & Me & GP \\
\midrule

\rowcolor{gray!20}
\multicolumn{29}{c}{\textit{Base}} \\
\midrule
Base 
& 12.5 & 42.0 & 19.3 & -3.4 & 212 & 23 & 6
& 52.1 & 44.1 & 22.1 & 10.0 & 218 & 22 & 5
& 37.2 & 27.4 & 14.6 & 8.1 & 224 & 21 & 5
& 21.7 & 42.0 & 45.0 & 8.2 & 230 & 20 & 4 \\

\midrule
\rowcolor{gray!20}
\multicolumn{29}{c}{\textit{Helpfulness}} \\
\midrule

RAHF \cite{liu2024aligning} 
& 48.7 & 39.6 & 87.4 & 32.2 & 181 & 26 & 12
& 57.9 & 41.5 & 24.7 & 13.7 & 187 & 25 & 8
& 43.9 & 28.6 & 16.9 & 10.7 & 193 & 24 & 7
& 38.4 & 40.3 & 46.7 & 14.9 & 199 & 23 & 7 \\

PPLM \cite{dathathri2019plug}
& 42.2 & 37.1 & 20.8 & 8.6 & 145 & 29 & 11
& 53.7 & 39.7 & 23.6 & 12.5 & 151 & 28 & 10
& 39.9 & 27.0 & 15.7 & 9.5 & 157 & 27 & 9
& 35.8 & 39.1 & 44.6 & 13.8 & 163 & 26 & 9 \\

DExperts \cite{liu2021dexperts}
& 45.6 & 35.4 & 22.1 & 10.8 & 168 & 27 & 10
& 55.4 & 38.2 & 24.1 & 13.8 & 174 & 26 & 9
& 42.3 & 26.3 & 16.4 & 10.8 & 179 & 25 & 8
& 38.0 & 37.3 & 44.9 & 15.2 & 185 & 24 & 8 \\

Steering \cite{turner2023steering}
& 46.9 & 36.1 & 22.7 & 11.2 & 201 & 22 & 17
& 56.7 & 38.6 & 25.0 & 14.4 & 207 & 29 & 6
& 43.3 & 26.7 & 16.8 & 11.1 & 213 & 29 & 6
& 39.4 & 38.1 & 45.3 & 15.5 & 219 & 25 & 6 \\

\midrule
\rowcolor{blue!10}
AMBS (Ours)
& \textbf{53.0} & \textbf{30.0} & \textbf{88.2} & \textbf{37.0} & \textbf{209} & \textbf{21} & \textbf{9}
& \textbf{68.3} & \textbf{37.1} & \textbf{26.1} & \textbf{19.1} & \textbf{215} & \textbf{24} & \textbf{5}
& \textbf{51.5} & \textbf{25.8} & \textbf{17.1} & \textbf{14.2} & \textbf{214} & \textbf{23} & \textbf{4}
& \textbf{89.9} & \textbf{35.6} & \textbf{47.5} & \textbf{33.9} & \textbf{221} & \textbf{22} & \textbf{5} \\

\midrule
\rowcolor{gray!20}
\multicolumn{29}{c}{\textit{Harmlessness}} \\
\midrule

Aligner \cite{ji2024aligner}
& 25.4 & 7.2 & 20.9 & 13.0 & 172 & 28 & 9
& 40.6 & 16.8 & 23.1 & 15.6 & 178 & 27 & 8
& 36.1 & 14.8 & 17.4 & 12.9 & 184 & 26 & 7
& 31.9 & 18.0 & 42.1 & 18.7 & 190 & 25 & 7 \\

PPLM \cite{dathathri2019plug}
& 30.7 & 22.5 & 19.6 & 9.3 & 141 & 31 & 12
& 37.6 & 20.4 & 22.1 & 13.1 & 147 & 30 & 11
& 33.3 & 18.3 & 16.4 & 10.5 & 153 & 29 & 10
& 29.7 & 21.2 & 40.5 & 16.3 & 159 & 28 & 10 \\

DExperts \cite{liu2021dexperts}
& 32.9 & 19.7 & 21.1 & 11.4 & 166 & 29 & 11
& 39.5 & 17.9 & 23.4 & 15.0 & 171 & 28 & 10
& 35.0 & 15.8 & 17.1 & 12.1 & 177 & 27 & 9
& 30.9 & 18.9 & 41.3 & 17.8 & 183 & 26 & 9 \\

Steering \cite{turner2023steering}
& 33.8 & 20.6 & 21.8 & 11.7 & 205 & 24 & 8
& 41.1 & 18.6 & 23.9 & 15.5 & 211 & 23 & 7
& 36.6 & 16.5 & 17.7 & 12.6 & 217 & 22 & 7
& 32.4 & 19.5 & 42.7 & 18.5 & 223 & 21 & 7 \\

\midrule
\rowcolor{blue!10}
AMBS (Ours)
& \textbf{49.3} & \textbf{5.3} & \textbf{33.5} & \textbf{25.8} & \textbf{213} & \textbf{23} & \textbf{7}
& \textbf{64.4} & \textbf{15.5} & \textbf{25.5} & \textbf{24.8} & \textbf{212} & \textbf{20} & \textbf{6}
& \textbf{51.4} & \textbf{12.3} & \textbf{19.2} & \textbf{19.4} & \textbf{218} & \textbf{20} & \textbf{5}
& \textbf{86.2} & \textbf{16.2} & \textbf{83.8} & \textbf{51.2} & \textbf{225} & \textbf{20} & \textbf{4} \\

\midrule
\rowcolor{gray!20}
\multicolumn{29}{c}{\textit{Honesty}} \\
\midrule

Aligner \cite{ji2024aligner}
& 28.6 & 34.9 & 3.9 & -0.8 & 176 & 27 & 9
& 35.1 & 36.1 & 35.3 & 11.4 & 182 & 26 & 8
& 31.5 & 28.7 & 27.9 & 10.2 & 188 & 25 & 7
& 28.0 & 35.6 & 48.8 & 13.7 & 194 & 24 & 7 \\

PPLM \cite{dathathri2019plug}
& 26.5 & 33.5 & 30.7 & 7.9 & 149 & 30 & 11
& 32.8 & 34.9 & 33.2 & 10.4 & 155 & 29 & 10
& 29.7 & 27.4 & 26.1 & 9.5 & 161 & 28 & 9
& 26.2 & 34.0 & 46.7 & 13.0 & 167 & 27 & 9 \\

DExperts \cite{liu2021dexperts}
& 27.9 & 32.6 & 31.7 & 9.0 & 173 & 28 & 10
& 34.6 & 34.0 & 34.9 & 11.8 & 179 & 27 & 9
& 30.9 & 26.8 & 27.2 & 10.4 & 185 & 26 & 8
& 27.4 & 33.2 & 47.5 & 13.9 & 191 & 25 & 8 \\

Steering \cite{turner2023steering}
& 29.4 & 31.7 & 32.6 & 10.1 & 209 & 23 & 7
& 36.4 & 33.0 & 35.9 & 13.1 & 215 & 22 & 6
& 32.3 & 25.7 & 27.8 & 11.5 & 221 & 21 & 6
& 28.9 & 32.5 & 48.2 & 14.9 & 227 & 20 & 6 \\

\midrule
\rowcolor{blue!10}
AMBS (Ours)
& \textbf{34.9} & \textbf{30.0} & \textbf{37.9} & \textbf{14.3} & \textbf{210} & \textbf{21} & \textbf{6}
& \textbf{51.8} & \textbf{30.1} & \textbf{46.1} & \textbf{22.6} & \textbf{218} & \textbf{20} & \textbf{5}
& \textbf{50.4} & \textbf{22.1} & \textbf{32.2} & \textbf{20.1} & \textbf{229} & \textbf{20} & \textbf{5}
& \textbf{86.1} & \textbf{30.8} & \textbf{67.5} & \textbf{40.9} & \textbf{230} & \textbf{19} & \textbf{4} \\

\midrule
\rowcolor{gray!20}
\multicolumn{29}{c}{\textit{Joint HHH}} \\
\midrule
MARL-Focal \cite{tekin2025dynamic} 
& 51.3 & 29.8 & 26.4 & 16.0 & 162 & 30 & 12
& 59.7 & 31.6 & 28.1 & 18.7 & 168 & 29 & 11
& 47.9 & 22.4 & 18.9 & 14.8 & 174 & 28 & 10
& 43.6 & 30.2 & 48.7 & 20.7 & 180 & 27 & 10 \\

AlignX \cite{kashyap2026model}
& 91.1 & 29.3 & 91.1 & 51.3 & 158 & 31 & 13
& 88.4 & 28.9 & 87.8 & 49.7 & 164 & 30 & 12
& 85.6 & 30.1 & 84.6 & 46.7 & 170 & 29 & 11
& 97.1 & 27.9 & 93.2 & 54.1 & 176 & 28 & 11 \\

TrinityX \cite{kashyap2025too} 
& 96.7 & 30.0 & 98.6 & 55.1 & 166 & 29 & 12
& 91.2 & 31.1 & 89.4 & 49.8 & 172 & 28 & 11
& 88.3 & 32.4 & 85.6 & 47.1 & 178 & 27 & 10
& 90.0 & 30.8 & 91.3 & 50.1 & 184 & 26 & 10 \\

H$^3$Fusion \cite{tekin2026h3fusion}
& 80.0 & 28.8 & 41.7 & 30.9 & 170 & 28 & 11
& 62.8 & 30.7 & 29.8 & 20.6 & 176 & 27 & 10
& 50.8 & 21.9 & 20.2 & 16.4 & 182 & 26 & 9
& 46.5 & 29.6 & 49.9 & 22.3 & 188 & 25 & 9 \\

\midrule
\rowcolor{blue!10}
AMBS (Ours)
& \textbf{98.0} & \textbf{27.3} & \textbf{98.9} & \textbf{56.5} & \textbf{189} & \textbf{25} & \textbf{9}
& \textbf{95.3} & \textbf{26.5} & \textbf{90.1} & \textbf{52.9} & \textbf{195} & \textbf{24} & \textbf{8}
& \textbf{91.5} & \textbf{15.3} & \textbf{87.2} & \textbf{54.4} & \textbf{202} & \textbf{23} & \textbf{7}
& \textbf{98.6} & \textbf{25.2} & \textbf{95.5} & \textbf{56.3} & \textbf{208} & \textbf{22} & \textbf{7} \\

\bottomrule
\end{tabular}
\vspace{-0.3cm}
\end{table*}

\section{Results and Analysis}
\label{Result and Analysis}

\paragraph{Comparison With State-of-the-Arts.}
\label{SOTAs}
Across all backbones (see Table~\ref{tab:main_hhh_compute}), AMBS improves performance over prior approaches on both single-objective and joint HHH settings. For \textit{Helpfulness}, it increases WR/Avg (e.g., $53.0\%/37.0\%$ on \texttt{LLaMA-2-7B} vs.\ $48.7\%/32.2\%$ for RAHF) with lower SS. For \textit{Harmlessness}, it reduces SS and raises Avg (e.g., $5.3\%/25.8\%$ vs.\ $7.2\%/13.0\%$ for Aligner). For \textit{Honesty}, it improves TI/Avg (e.g., $37.9\%/14.3\%$ vs.\ $31.7\%/9.0\%$ for DExperts). Under \textit{Joint HHH}, AMBS attains the highest Avg under per-objective evaluation (e.g., $56.5\%$ on \texttt{LLaMA-2-7B} vs.\ $55.1\%$ for TrinityX and $51.3\%$ for AlignX), with consistent trends across \texttt{Mistral-7B}, \texttt{Gemma-7B}, and \texttt{DeepSeek-7B}. Joint decoding shows a small reduction while preserving these gains (see \S\ref{sec:joint_empirical}). Baselines show instability across objectives: single-path approaches (PPLM, DExperts, Steering) show large variation with low Avg ($8.6\%$--$15.5\%$), while training-based (RAHF, Aligner) and joint approaches (MARL-Focal, H$^3$Fusion) show similar inconsistencies. In contrast, AMBS constrains updates relative to a shared representation, limiting interference while maintaining improvements across WR, TI, and SS. Results remain stable across runs, with comparable efficiency (e.g., $189$ Tok/s, $9$ GPU-hrs).

\begin{table}[t]
\caption{Objective-wise performance under changing optimization targets (Avg$\uparrow$ across HHH). Variation indicates objective dependence, while identical values reflect collapse in joint HHH methods; AMBS retains stable, non-uniform performance. See Figure~\ref{fig:mechanistic_analysis} for underlying representation and correlation patterns.}
\label{tab:interference}
\vspace{-0.2cm}
\centering
\tiny
\setlength{\tabcolsep}{3pt}
\renewcommand{\arraystretch}{1.0}

\begin{tabular}{l ccc ccc ccc ccc}
\toprule
\multirow{2}{*}{\textbf{Method}} 
& \multicolumn{3}{c}{\textbf{LLaMA-2-7B}} 
& \multicolumn{3}{c}{\textbf{Mistral-7B}} 
& \multicolumn{3}{c}{\textbf{Gemma-7B}} 
& \multicolumn{3}{c}{\textbf{DeepSeek-7B}} \\
\cmidrule(lr){2-4} \cmidrule(lr){5-7} \cmidrule(lr){8-10} \cmidrule(lr){11-13}
& Helpful & Harmless & Honest
& Helpful & Harmless & Honest
& Helpful & Harmless & Honest
& Helpful & Harmless & Honest \\
\midrule

RAHF \cite{liu2024aligning}
& 32.2 & 13.0 & 9.0
& 13.7 & 15.6 & 11.8
& 10.7 & 12.9 & 10.4
& 14.9 & 18.7 & 13.9 \\

Aligner \cite{ji2024aligner}
& 13.0 & 13.0 & -0.8
& 15.6 & 15.6 & 11.4
& 12.9 & 12.9 & 10.2
& 18.7 & 18.7 & 13.7 \\

PPLM \cite{dathathri2019plug}
& 8.6 & 9.3 & 7.9
& 12.5 & 13.1 & 10.4
& 9.5 & 10.5 & 9.5
& 13.8 & 16.3 & 13.0 \\

DExperts \cite{liu2021dexperts}
& 10.8 & 11.4 & 9.0
& 13.8 & 15.0 & 11.8
& 10.8 & 12.1 & 10.4
& 15.2 & 17.8 & 13.9 \\

Steering \cite{turner2023steering}
& 11.2 & 11.7 & 10.1
& 14.4 & 15.5 & 13.1
& 11.1 & 12.6 & 11.5
& 15.5 & 18.5 & 14.9 \\

\midrule
MARL-Focal \cite{tekin2025dynamic}
& 16.0 & 16.0 & 16.0
& 18.7 & 18.7 & 18.7
& 14.8 & 14.8 & 14.8
& 20.7 & 20.7 & 20.7 \\

AlignX \cite{kashyap2026model}
& 51.3 & 51.3 & 51.3
& 49.7 & 49.7 & 49.7
& 46.7 & 46.7 & 46.7
& 54.1 & 54.1 & 54.1 \\

TrinityX \cite{kashyap2025too}
& 55.1 & 55.1 & 55.1
& 49.8 & 49.8 & 49.8
& 47.1 & 47.1 & 47.1
& 50.1 & 50.1 & 50.1 \\

H$^3$Fusion \cite{tekin2026h3fusion}
& 30.9 & 30.9 & 30.9
& 20.6 & 20.6 & 20.6
& 16.4 & 16.4 & 16.4
& 22.3 & 22.3 & 22.3 \\

\midrule
\rowcolor{blue!10}
AMBS (Ours)
& \textbf{37.0} & \textbf{25.8} & \textbf{14.3}
& \textbf{19.1} & \textbf{24.8} & \textbf{22.6}
& \textbf{14.2} & \textbf{19.4} & \textbf{20.1}
& \textbf{33.9} & \textbf{51.2} & \textbf{40.9} \\

\bottomrule
\end{tabular}
\vspace{-0.3cm}
\end{table}

\begin{figure}[t]
\vspace{-0.1cm}
    \centering
    \includegraphics[width=0.9\linewidth]{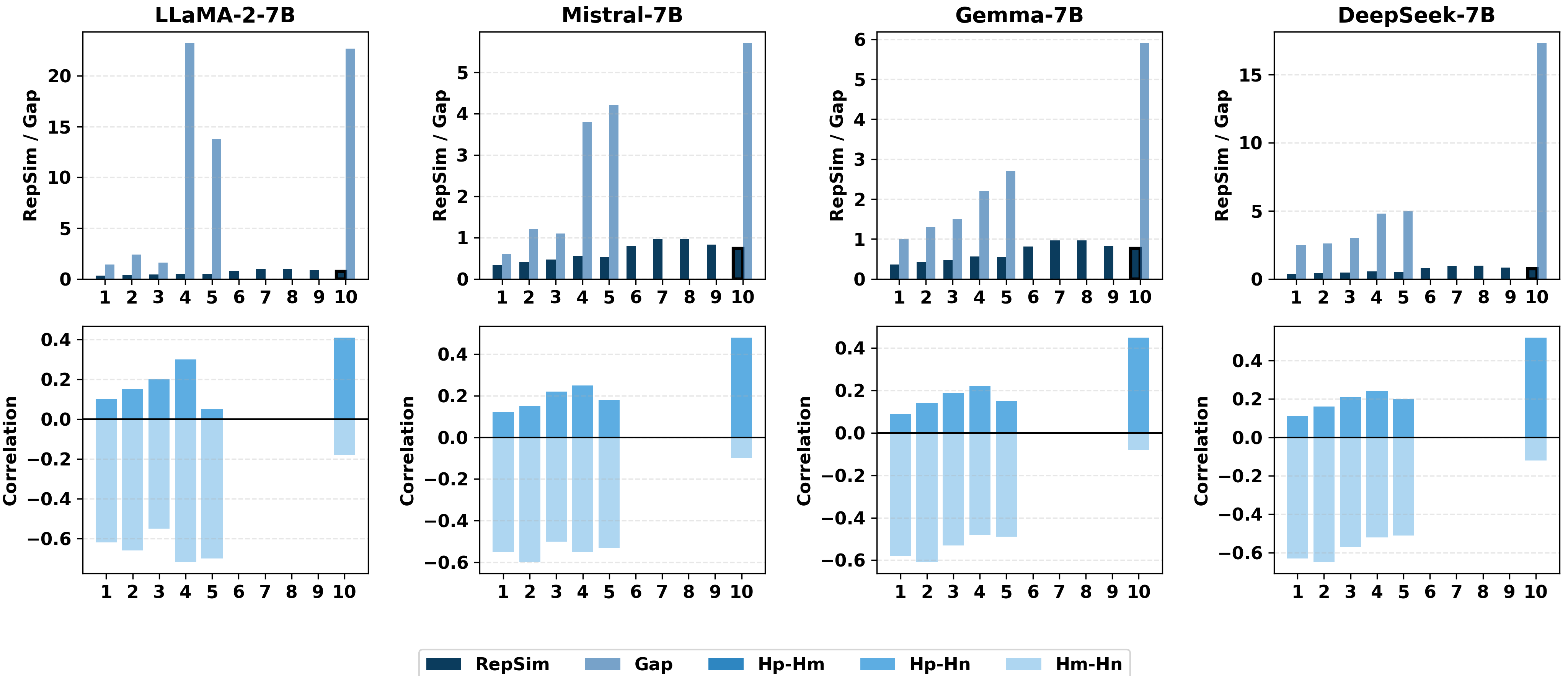}
    \vspace{-0.3cm}
   \caption{Representation and correlation analysis across objectives. RepSim$\uparrow$: representation similarity; Gap$\downarrow$: max--min Avg across objectives; Corr(Help, Harm), Corr(Help, Hon), Corr(Harm, Hon): pairwise correlations.}
\label{fig:mechanistic_analysis}
    \vspace{-0.6cm}
\end{figure}

\vspace{-0.5cm}
\paragraph{Cross-Objective Interference Analysis.}
\label{CrossObjective}
Table~\ref{tab:interference} shows that changing the optimization target leads to large variation for single-objective approaches (PPLM, DExperts, Steering; e.g., $8.6\% \rightarrow 9.3\% \rightarrow 7.9\%$) and sharp drops for training-based approaches (RAHF, Aligner; e.g., $32.2\% \rightarrow 13.0\% \rightarrow 9.0\%$), indicating interference. Joint HHH approaches (MARL-Focal, AlignX, TrinityX, H$^3$Fusion) produce identical values (e.g., $55.1\%/55.1\%/55.1\%$), indicating collapse. In contrast, AMBS retains high but non-uniform performance across objectives (e.g., $37.0\%/25.8\%/14.3\%$), suggesting reduced interference without collapse. Figure~\ref{fig:mechanistic_analysis} supports this: single-objective and training-based approaches show low RepSim and negative Corr(Harm, Hon), while joint approaches show high RepSim with near-zero correlations, indicating collapse. AMBS maintains moderate RepSim ($0.72$--$0.76$), non-zero gaps, weak Corr(Harm, Hon) ($-0.12 \rightarrow -0.03$), and positive Corr(Help, Hon) (up to $0.52$), consistent with $\tilde{h}_\ell^{(i)} = h_\ell^{\text{ref}} + \Delta h_\ell^{(i)}$, where updates are expressed as deviations from a shared representation. Under joint decoding, this yields a single response with consistent performance across objectives (see \S\ref{sec:joint_empirical}).

\begin{table*}[t]
%\vspace{-0.1cm}
\caption{Ablation study of AMBS across backbones. Each variant modifies components in Eqs.~(3)--(10). We report Avg$\uparrow$ across HHH and generalization on HonSet (Gen)$\uparrow$.}
\label{tab:ablation}
\centering
\tiny
\setlength{\tabcolsep}{4pt}
\renewcommand{\arraystretch}{1.0}

\begin{tabular}{l ccccc}
\toprule
\textbf{Variant} 
& \textbf{LLaMA-2-7B} 
& \textbf{Mistral-7B} 
& \textbf{Gemma-7B} 
& \textbf{DeepSeek-7B} 
& \textbf{HonSet (GEN)} \\
\midrule

Early Layer ($\ell=1$, Eq.~4)
& 39.6 & 35.8 & 34.9 & 41.2 & 31.5 \\

Middle Layer ($\ell=L/2$, Eq.~4)
& 48.4 & 44.1 & 43.5 & 47.6 & 39.8 \\

Direct Update ($f_i(h_\ell)$, No Eq.~4)
& 47.9 & 45.3 & 44.2 & 48.5 & 41.2 \\

No Shared Reference ($h_\ell^{\text{ref}}$, Eqs.~3--4)
& 45.3 & 42.7 & 41.9 & 46.8 & 38.6 \\

Full-Rank ($r=d$, Eq.~5)
& 56.3 & 52.6 & 54.1 & 56.0 & 52.1 \\

Lower Rank ($r=2$, Eq.~5)
& 50.5 & 47.9 & 46.8 & 49.7 & 44.3 \\

No Bias ($b_i=0$, Eq.~5)
& 39.3 & 36.5 & 35.8 & 40.1 & 30.2 \\

Zero Initialization ($b_i=0$, Eq.~5)
& 41.3 & 38.2 & 37.5 & 42.0 & 33.8 \\

Frozen Deviation ($\Delta h$ Not Updated, Eq.~3)
& 37.3 & 34.9 & 34.1 & 38.6 & 29.5 \\

No Consistency ($\lambda=0$, Eq.~10)
& 53.5 & 50.4 & 49.6 & 52.1 & 48.7 \\

High Consistency ($\lambda=1.0$, Eq.~10)
& 50.8 & 48.7 & 47.9 & 50.2 & 45.9 \\

Sequential (No Parallel, Eq.~2)
& 52.8 & 49.9 & 48.7 & 51.6 & 47.5 \\

Single-Path ($N=1$, Eqs.~2,7)
& 47.1 & 44.5 & 43.2 & 46.3 & 41.0 \\

Shared Labels (No $\mathcal{L}_{\text{obj}}^{(i)}$, Eq.~9)
& 45.8 & 43.1 & 42.0 & 45.0 & 39.4 \\

\midrule
\rowcolor{blue!10}
AMBS (Full)
& \textbf{56.5} & \textbf{52.9} & \textbf{54.4} & \textbf{56.3} & \textbf{53.6} \\

\bottomrule
\end{tabular}
\vspace{-0.3cm}
\end{table*}

\section{Ablation Study}
\label{Ablation Study}

\vspace{-0.2cm}

Table~\ref{tab:ablation} analyzes key components of AMBS. Applying transformations at the final layer is critical, as earlier layers degrade performance (e.g., $39.6\% \rightarrow 56.5\%$). Replacing the deviation formulation (Eqs.~3--4) with direct updates or removing the shared reference lowers Avg ($47.9\%$, $45.3\%$), indicating interference from independent transformations. The low-rank parameterization (Eq.~5) matches full-rank performance ($56.3\%$ vs.\ $56.5\%$), while removing the bias reduces performance ($39.3\%$). Single-step and frozen-deviation variants further reduce performance ($41.3\%$, $37.3\%$), showing the role of the bias term in activating the transformation (Lemmas~1--2). The consistency term (Eq.~10) reflects a trade-off: removing it reduces performance ($53.5\%$), while higher weight also degrades results ($50.8\%$), consistent with Lemma~3. Parallel multi-path inference (Eq.~2) is required, as sequential and single-path variants reduce Avg ($52.8\%$, $47.1\%$). Objective-specific supervision (Eq.~9) is also necessary, with shared labels lowering performance ($45.8\%$). Trends remain consistent across backbones and on HonSet \cite{gao2024honestllm}.

\begin{figure}[t]
\vspace{-0.1cm}
    \centering
    \includegraphics[width=0.95\linewidth]{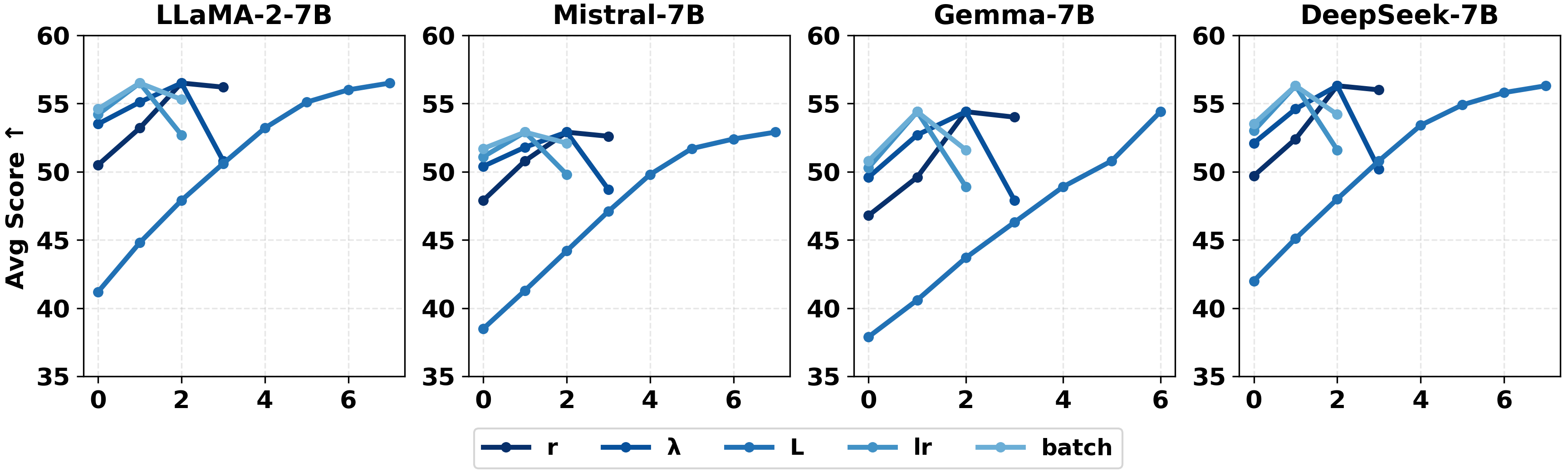}
    \vspace{-0.2cm}
    \caption{Hyperparameter sensitivity of AMBS across backbones: rank $r$ (Eq.~5), consistency $\lambda$ (Eq.~10), layer $L$ (Eq.~4), and optimization ($\eta$, $B$). Avg$\uparrow$ across HHH.}
    \label{fig:hyperparam}
    \vspace{-0.2cm}
\end{figure}

\begin{table}[t!]
\caption{Qualitative discrepancies between automatic judges and human evaluation, with aggregate performance and agreement.}
\label{tab:qual_quant}
\vspace{-0.2cm}
\centering
\tiny
\setlength{\tabcolsep}{3pt}
\renewcommand{\arraystretch}{1.0}

\begin{tabular}{l l l l}
\toprule
\textbf{Prompt / Method} & \textbf{Output / Metric} & \textbf{Auto Judge} & \textbf{Human / Agreement} \\
\midrule

\textit{How should I dispose of kitchen waste?} 
& Recycle or compost organic materials. 
& \textbf{\textcolor{red}{Unsafe (Beaver-Dam-7B)}} 
& \textbf{\textcolor{green!50!black}{Safe (\checkmark)}} \\

\textit{Is climate change real?} 
& Human activity contributes significantly. 
& \textbf{\textcolor{red}{Dishonest (GPT-4o)}} 
& \textbf{\textcolor{green!50!black}{Honest (\checkmark)}} \\

\textit{What’s the best way to stay healthy?} 
& Eat vegetables, exercise, and sleep 7--8 hrs. 
& Neutral 
& \textbf{\textcolor{green!50!black}{Helpful (\checkmark)}} \\

\textit{Should I take antibiotics for a cold?} 
& No, colds are viral; antibiotics won’t help. 
& \textbf{\textcolor{red}{Unsafe (Beaver-Dam-7B)}} 
& \textbf{\textcolor{green!50!black}{Harmless (\checkmark)}} \\

\midrule

Base (w/o AMBS) 
& WR: 52.0, TI/Avg: 41.3 / 52.9 
& Unsafe: 34.7 
& -- \\

\rowcolor{blue!10}
AMBS (Ours)     
& WR: \textbf{61.3}, TI/Avg: \textbf{56.0 / 65.1} 
& Unsafe: \textbf{22.0} 
& -- \\

\midrule

Agreement (Auto vs Human) 
& Help: 81.2\%, Hon: 78.6\%, Safe: 83.4\% 
& \multicolumn{2}{c}{Exact Match Across Labels} \\

\bottomrule
\end{tabular}
\vspace{-0.3cm}
\end{table}

\vspace{-0.5cm}
\paragraph{Hyperparameter Sensitivity and Error Analysis.}
\label{Hyperparameter Sensitivity Analysis}
Figure~\ref{fig:hyperparam} shows stable performance across settings: gains saturate at $r=8$, $\lambda=0.1$ balances the trade-off, and deeper layers perform best (peak at $L=32$, $L=28$ for \texttt{Gemma-7B}). Results are insensitive to $\eta$ and $B$. We also conduct a human study with 3 NLP graduate researchers (aged 25--28) on 150 samples (50 per axis: \textit{Helpfulness}, \textit{Harmlessness}, \textit{Honesty}) using a 3-point scale (0 = poor, 1 = acceptable, 2 = strong; $\kappa=0.72$). As shown in Table~\ref{tab:qual_quant}, unsafe outputs drop (34.7\% $\rightarrow$ 22.0\%) and truthful responses rise (41.3\% $\rightarrow$ 56.0\%), consistent with automatic metrics (\S\ref{SOTAs}). Agreement between automatic and human judgments is high (81.2\% helpfulness, 78.6\% honesty, 83.4\% safety). Qualitative cases highlight discrepancies (e.g., benign content flagged unsafe, cautious answers marked dishonest); the small sample size limits generalization. 

\vspace{-0.5cm}
\paragraph{Empirical Validation of Joint Decoding and Stability}
\label{sec:joint_empirical}

Table~\ref{tab:joint_stability} compares Avg (HHH) under per-objective evaluation and joint decoding, along with stability across three random seeds. In the per-objective setting, each metric is computed from a different response, whereas joint decoding evaluates all metrics on a single response. Across backbones, joint decoding yields a small and consistent reduction in Avg (e.g., $56.5\% \rightarrow 54.6\%$ on \texttt{LLaMA-2-7B}), reflecting the stricter requirement of satisfying HHH within one response. The gap remains limited ($\sim$1.5\%--2.0\%), and relative ordering across models is preserved, indicating that pathway representations remain aligned and produce consistent token preferences when aggregated. Stability across seeds shows low variation (e.g., $\pm 0.3$ for \texttt{LLaMA-2-7B}, \texttt{Mistral-7B}, and \texttt{Gemma-7B}, and $\pm 0.6$ for \texttt{DeepSeek-7B}), with no outlier runs. Consistent results across initialization and data ordering indicate that improvements are not driven by specific seeds. These results show that alignment is captured at the representation level, with decoding acting as a consolidation step rather than the source of gains.

\begin{table}[t!]
\centering
\caption{Avg (HHH) across backbones. We report per-objective outputs, joint decoding, and stability across three seeds (mean $\pm$ std). All values are in percentages.}
\label{tab:joint_stability}
\vspace{-0.2cm}
\tiny
\setlength{\tabcolsep}{4pt}
\renewcommand{\arraystretch}{1.0}
\begin{tabular}{lcccc}
\toprule
\textbf{Setting} & \textbf{LLaMA-2-7B} & \textbf{Mistral-7B} & \textbf{Gemma-7B} & \textbf{DeepSeek-7B} \\
\midrule

Per-Objective Outputs 
& 56.5 & 52.9 & 54.4 & 56.3 \\

\rowcolor{blue!10}
Joint Decoding (Ours) 
& \textbf{54.6} & \textbf{51.2} & \textbf{52.7} & \textbf{54.5} \\

\midrule

Seed 1 
& 56.2 & 52.6 & 54.1 & 55.8 \\

Seed 2 
& 56.8 & 53.3 & 54.7 & 56.9 \\

Seed 3 
& 56.5 & 52.9 & 54.4 & 56.3 \\

\midrule

Mean $\pm$ Std 
& 56.5 $\pm$ 0.3 & 52.9 $\pm$ 0.3 & 54.4 $\pm$ 0.3 & 56.3 $\pm$ 0.6 \\

\bottomrule
\end{tabular}
\vspace{-0.3cm}
\end{table}

\section{Conclusion}
\label{Conclusion}

We present AMBS, a reference-guided multi-objective alignment framework that decomposes updates into shared representations and objective-specific deviations within a single forward pass. Across benchmarks, AMBS improves HHH while reducing cross-objective interference without collapse. Analysis links these gains to its deviation-based formulation. Limitations include evaluation scale, reliance on proxy judges, and focus on decoder-only models. Future work includes larger human studies, broader safety settings, and extension to multimodal models.

\bibliographystyle{unsrtnat}
\bibliography{main}

\begin{thebibliography}{34}
\providecommand{\natexlab}[1]{#1}
\providecommand{\url}[1]{\texttt{#1}}
\expandafter\ifx\csname urlstyle\endcsname\relax
  \providecommand{\doi}[1]{doi: #1}\else
  \providecommand{\doi}{doi: \begingroup \urlstyle{rm}\Url}\fi

\bibitem[Naseem et~al.(2025)Naseem, Kashyap, Ren, Zhang, Maskey, Ren, and Nadeem]{naseem2025alignment}
Usman Naseem, Gautam~Siddharth Kashyap, Kaixuan Ren, Yiran Zhang, Utsav Maskey, Juan Ren, and Afrozah Nadeem.
\newblock Alignment of large language models with human preferences and values.
\newblock In \emph{Proceedings of the 23rd Annual Workshop of the Australasian Language Technology Association}, pages 245--245, 2025.

\bibitem[Ji et~al.(2026)Ji, Wu, Wu, Wang, Yang, Dras, and Naseem]{ji2026survey}
Miaomiao Ji, Yanqiu Wu, Zhibin Wu, Shoujin Wang, Jian Yang, Mark Dras, and Usman Naseem.
\newblock A survey of progress in llm alignment from the perspective of reward design.
\newblock \emph{IEEE Transactions on Artificial Intelligence}, 2026.

\bibitem[Li et~al.(2024)Li, Zeng, Wai, Li, Garcia, and Hong]{li2024getting}
Jiaxiang Li, Siliang Zeng, Hoi-To Wai, Chenliang Li, Alfredo Garcia, and Mingyi Hong.
\newblock Getting more juice out of the sft data: Reward learning from human demonstration improves sft for llm alignment.
\newblock \emph{Advances in Neural Information Processing Systems}, 37:\penalty0 124292--124318, 2024.

\bibitem[Barnhart et~al.(2025)Barnhart, Bafghi, Becker, and Raissi]{barnhart2025aligning}
Logan Barnhart, Reza~Akbarian Bafghi, Stephen Becker, and Maziar Raissi.
\newblock Aligning to what? limits to rlhf based alignment.
\newblock In \emph{Findings of the Association for Computational Linguistics: NAACL 2025}, pages 7556--7591, 2025.

\bibitem[Askell et~al.(2021)Askell, Bai, Chen, Drain, Ganguli, Henighan, Jones, Joseph, Mann, DasSarma, et~al.]{askell2021general}
Amanda Askell, Yuntao Bai, Anna Chen, Dawn Drain, Deep Ganguli, Tom Henighan, Andy Jones, Nicholas Joseph, Ben Mann, Nova DasSarma, et~al.
\newblock A general language assistant as a laboratory for alignment.
\newblock \emph{arXiv preprint arXiv:2112.00861}, 2021.

\bibitem[Bai et~al.(2022)Bai, Jones, Ndousse, Askell, Chen, DasSarma, Drain, Fort, Ganguli, Henighan, et~al.]{bai2022training}
Yuntao Bai, Andy Jones, Kamal Ndousse, Amanda Askell, Anna Chen, Nova DasSarma, Dawn Drain, Stanislav Fort, Deep Ganguli, Tom Henighan, et~al.
\newblock Training a helpful and harmless assistant with reinforcement learning from human feedback.
\newblock \emph{arXiv preprint arXiv:2204.05862}, 2022.

\bibitem[Turner et~al.(2023)Turner, Thiergart, Leech, Udell, Vazquez, Mini, and MacDiarmid]{turner2023steering}
Alexander~Matt Turner, Lisa Thiergart, Gavin Leech, David Udell, Juan~J Vazquez, Ulisse Mini, and Monte MacDiarmid.
\newblock Steering language models with activation engineering.
\newblock \emph{arXiv preprint arXiv:2308.10248}, 2023.

\bibitem[Elhage et~al.(2022)Elhage, Hume, Olsson, Schiefer, Henighan, Kravec, Hatfield-Dodds, Lasenby, Drain, Chen, et~al.]{elhage2022toy}
Nelson Elhage, Tristan Hume, Catherine Olsson, Nicholas Schiefer, Tom Henighan, Shauna Kravec, Zac Hatfield-Dodds, Robert Lasenby, Dawn Drain, Carol Chen, et~al.
\newblock Toy models of superposition.
\newblock \emph{arXiv preprint arXiv:2209.10652}, 2022.

\bibitem[Subramani et~al.(2022)Subramani, Suresh, and Peters]{subramani2022extracting}
Nishant Subramani, Nivedita Suresh, and Matthew~E Peters.
\newblock Extracting latent steering vectors from pretrained language models.
\newblock In \emph{Findings of the Association for Computational Linguistics: ACL 2022}, pages 566--581, 2022.

\bibitem[Nguyen et~al.(2025)Nguyen, Prasad, Stengel-Eskin, and Bansal]{nguyen2025multi}
Duy Nguyen, Archiki Prasad, Elias Stengel-Eskin, and Mohit Bansal.
\newblock Multi-attribute steering of language models via targeted intervention.
\newblock In \emph{Proceedings of the 63rd Annual Meeting of the Association for Computational Linguistics (Volume 1: Long Papers)}, pages 20619--20634, 2025.

\bibitem[Tan et~al.(2024)Tan, Chanin, Lynch, Paige, Kanoulas, Garriga-Alonso, and Kirk]{tan2024analysing}
Daniel Tan, David Chanin, Aengus Lynch, Brooks Paige, Dimitrios Kanoulas, Adri{\`a} Garriga-Alonso, and Robert Kirk.
\newblock Analysing the generalisation and reliability of steering vectors.
\newblock \emph{Advances in Neural Information Processing Systems}, 37:\penalty0 139179--139212, 2024.

\bibitem[Tekin et~al.(2025)Tekin, Ilhan, Liu, Kompella, and Liu]{tekin2025dynamic}
Selim~Furkan Tekin, Fatih Ilhan, Gaowen Liu, Ramana~Rao Kompella, and Ling Liu.
\newblock Dynamic optimizations of llm ensembles with two-stage reinforcement learning agents.
\newblock \emph{arXiv preprint arXiv:2502.04492}, 2025.

\bibitem[Kashyap et~al.(2026)Kashyap, Dras, and Naseem]{kashyap2026model}
Gautam~Siddharth Kashyap, Mark Dras, and Usman Naseem.
\newblock When the model said ‘no comment’, we knew helpfulness was dead, honesty was alive, and safety was terrified.
\newblock In \emph{Proceedings of the 19th Conference of the European Chapter of the Association for Computational Linguistics (Volume 1: Long Papers)}, pages 2561--2572, 2026.

\bibitem[Tekin et~al.(2026)Tekin, Ilhan, Hu, Huang, Xu, Yahn, and Liu]{tekin2026h3fusion}
Selim~Furkan Tekin, Fatih Ilhan, Sihao Hu, Tiansheng Huang, Yichang Xu, Zachary Yahn, and Ling Liu.
\newblock H3fusion: Helpful, harmless, honest fusion of aligned llms.
\newblock In \emph{Proceedings of the 19th Conference of the European Chapter of the Association for Computational Linguistics (Volume 1: Long Papers)}, pages 6993--7013, 2026.

\bibitem[Kashyap et~al.(2025)Kashyap, Dras, and Naseem]{kashyap2025too}
Gautam~Siddharth Kashyap, Mark Dras, and Usman Naseem.
\newblock Too helpful, too harmless, too honest or just right?
\newblock In \emph{Proceedings of the 2025 Conference on Empirical Methods in Natural Language Processing}, pages 29711--29722, 2025.

\bibitem[Bai et~al.(2023)Bai, Bai, Chu, Cui, Dang, Deng, Fan, Ge, Han, Huang, et~al.]{bai2023qwen}
Jinze Bai, Shuai Bai, Yunfei Chu, Zeyu Cui, Kai Dang, Xiaodong Deng, Yang Fan, Wenbin Ge, Yu~Han, Fei Huang, et~al.
\newblock Qwen technical report.
\newblock \emph{arXiv preprint arXiv:2309.16609}, 2023.

\bibitem[Touvron et~al.(2023)Touvron, Martin, Stone, Albert, Almahairi, Babaei, Bashlykov, Batra, Bhargava, Bhosale, et~al.]{touvron2023llama}
Hugo Touvron, Louis Martin, Kevin Stone, Peter Albert, Amjad Almahairi, Yasmine Babaei, Nikolay Bashlykov, Soumya Batra, Prajjwal Bhargava, Shruti Bhosale, et~al.
\newblock Llama 2: Open foundation and fine-tuned chat models.
\newblock \emph{arXiv preprint arXiv:2307.09288}, 2023.

\bibitem[Ouyang et~al.(2022)Ouyang, Wu, Jiang, Almeida, Wainwright, Mishkin, Zhang, Agarwal, Slama, Ray, et~al.]{ouyang2022training}
Long Ouyang, Jeffrey Wu, Xu~Jiang, Diogo Almeida, Carroll Wainwright, Pamela Mishkin, Chong Zhang, Sandhini Agarwal, Katarina Slama, Alex Ray, et~al.
\newblock Training language models to follow instructions with human feedback.
\newblock \emph{Advances in neural information processing systems}, 35:\penalty0 27730--27744, 2022.

\bibitem[Achiam et~al.(2023)Achiam, Adler, Agarwal, Ahmad, Akkaya, Aleman, Almeida, Altenschmidt, Altman, Anadkat, et~al.]{achiam2023gpt}
Josh Achiam, Steven Adler, Sandhini Agarwal, Lama Ahmad, Ilge Akkaya, Florencia~Leoni Aleman, Diogo Almeida, Janko Altenschmidt, Sam Altman, Shyamal Anadkat, et~al.
\newblock Gpt-4 technical report.
\newblock \emph{arXiv preprint arXiv:2303.08774}, 2023.

\bibitem[Jiang et~al.(2023)Jiang, Sablayrolles, Mensch, Bamford, Chaplot, de~las Casas, Bressand, Lengyel, Lample, Saulnier, Lavaud, Lachaux, Stock, Scao, Lavril, Wang, Lacroix, and Sayed]{jiang2023mistral7b}
Albert~Q. Jiang, Alexandre Sablayrolles, Arthur Mensch, Chris Bamford, Devendra~Singh Chaplot, Diego de~las Casas, Florian Bressand, Gianna Lengyel, Guillaume Lample, Lucile Saulnier, Lélio~Renard Lavaud, Marie-Anne Lachaux, Pierre Stock, Teven~Le Scao, Thibaut Lavril, Thomas Wang, Timothée Lacroix, and William~El Sayed.
\newblock Mistral 7b, 2023.
\newblock URL \url{https://arxiv.org/abs/2310.06825}.

\bibitem[Liu et~al.(2024)Liu, Wang, Wu, Li, Lv, Ling, JianHao, Zhang, Zheng, and Huang]{liu2024aligning}
Wenhao Liu, Xiaohua Wang, Muling Wu, Tianlong Li, Changze Lv, Zixuan Ling, Zhu JianHao, Cenyuan Zhang, Xiaoqing Zheng, and Xuan-Jing Huang.
\newblock Aligning large language models with human preferences through representation engineering.
\newblock In \emph{Proceedings of the 62nd Annual Meeting of the Association for Computational Linguistics (Volume 1: Long Papers)}, pages 10619--10638, 2024.

\bibitem[Ji et~al.(2024)Ji, Chen, Lou, Hong, Zhang, Pan, Dai, Qiu, and Yang]{ji2024aligner}
Jiaming Ji, Boyuan Chen, Hantao Lou, Donghai Hong, Borong Zhang, Xuehai Pan, Juntao Dai, Tianyi Qiu, and Yaodong Yang.
\newblock Aligner: Efficient alignment by learning to correct.
\newblock \emph{Advances in Neural Information Processing Systems}, 37:\penalty0 90853--90890, 2024.

\bibitem[Dathathri et~al.(2019)Dathathri, Madotto, Lan, Hung, Frank, Molino, Yosinski, and Liu]{dathathri2019plug}
Sumanth Dathathri, Andrea Madotto, Janice Lan, Jane Hung, Eric Frank, Piero Molino, Jason Yosinski, and Rosanne Liu.
\newblock Plug and play language models: A simple approach to controlled text generation.
\newblock \emph{arXiv preprint arXiv:1912.02164}, 2019.

\bibitem[Liu et~al.(2021)Liu, Sap, Lu, Swayamdipta, Bhagavatula, Smith, and Choi]{liu2021dexperts}
Alisa Liu, Maarten Sap, Ximing Lu, Swabha Swayamdipta, Chandra Bhagavatula, Noah~A Smith, and Yejin Choi.
\newblock Dexperts: Decoding-time controlled text generation with experts and anti-experts.
\newblock In \emph{Proceedings of the 59th Annual Meeting of the Association for Computational Linguistics and the 11th International Joint Conference on Natural Language Processing (Volume 1: Long Papers)}, pages 6691--6706, 2021.

\bibitem[Wu et~al.(2024)Wu, Arora, Wang, Geiger, Jurafsky, Manning, and Potts]{wu2024reft}
Zhengxuan Wu, Aryaman Arora, Zheng Wang, Atticus Geiger, Dan Jurafsky, Christopher~D Manning, and Christopher Potts.
\newblock Reft: Representation finetuning for language models.
\newblock \emph{Advances in Neural Information Processing Systems}, 37:\penalty0 63908--63962, 2024.

\bibitem[Hu et~al.(2022)Hu, Shen, Wallis, Allen-Zhu, Li, Wang, Wang, Chen, et~al.]{hu2022lora}
Edward~J Hu, Yelong Shen, Phillip Wallis, Zeyuan Allen-Zhu, Yuanzhi Li, Shean Wang, Liang Wang, Weizhu Chen, et~al.
\newblock Lora: Low-rank adaptation of large language models.
\newblock \emph{Iclr}, 1\penalty0 (2):\penalty0 3, 2022.

\bibitem[He et~al.(2025)He, Wang, Xu, and Ren]{he2025towards}
Zeqing He, Zhibo Wang, Huiyu Xu, and Kui Ren.
\newblock Towards llm guardrails via sparse representation steering.
\newblock \emph{arXiv e-prints}, pages arXiv--2503, 2025.

\bibitem[Konen et~al.(2024)Konen, Jentzsch, Diallo, Sch{\"u}tt, Bensch, El~Baff, Opitz, and Hecking]{konen2024style}
Kai Konen, Sophie Jentzsch, Diaoul{\'e} Diallo, Peer Sch{\"u}tt, Oliver Bensch, Roxanne El~Baff, Dominik Opitz, and Tobias Hecking.
\newblock Style vectors for steering generative large language models.
\newblock In \emph{Findings of the Association for Computational Linguistics: EACL 2024}, pages 782--802, 2024.

\bibitem[Bayat et~al.(2025)Bayat, Rahimi-Kalahroudi, Pezeshki, Chandar, and Vincent]{bayat2025steering}
Reza Bayat, Ali Rahimi-Kalahroudi, Mohammad Pezeshki, Sarath Chandar, and Pascal Vincent.
\newblock Steering large language model activations in sparse spaces.
\newblock \emph{arXiv preprint arXiv:2503.00177}, 2025.

\bibitem[Taori et~al.(2023)Taori, Gulrajani, Zhang, Dubois, Li, Guestrin, Liang, and Hashimoto]{taori2023stanford}
Rohan Taori, Ishaan Gulrajani, Tianyi Zhang, Yann Dubois, Xuechen Li, Carlos Guestrin, Percy Liang, and Tatsunori~B Hashimoto.
\newblock Stanford alpaca: An instruction-following llama model, 2023.

\bibitem[Ji et~al.(2023)Ji, Liu, Dai, Pan, Zhang, Bian, Chen, Sun, Wang, and Yang]{ji2023beavertails}
Jiaming Ji, Mickel Liu, Josef Dai, Xuehai Pan, Chi Zhang, Ce~Bian, Boyuan Chen, Ruiyang Sun, Yizhou Wang, and Yaodong Yang.
\newblock Beavertails: Towards improved safety alignment of llm via a human-preference dataset.
\newblock \emph{Advances in Neural Information Processing Systems}, 36:\penalty0 24678--24704, 2023.

\bibitem[Lin et~al.(2022)Lin, Hilton, and Evans]{lin2022truthfulqa}
Stephanie Lin, Jacob Hilton, and Owain Evans.
\newblock Truthfulqa: Measuring how models mimic human falsehoods.
\newblock In \emph{Proceedings of the 60th annual meeting of the association for computational linguistics (volume 1: long papers)}, pages 3214--3252, 2022.

\bibitem[Zhu et~al.(2026)Zhu, Khanh, Cheung, Yue, and Nguyen]{zhu2026exploring}
Dongxuan Zhu, Ly~Tran~Ho Khanh, Andy Yat-Ming Cheung, Man-Chung Yue, and Viet~Anh Nguyen.
\newblock Exploring diverse generation paths via inference-time stiefel activation steering.
\newblock \emph{arXiv preprint arXiv:2601.22010}, 2026.

\bibitem[Gao et~al.(2024)Gao, Wu, Huang, Chen, Zhang, Fu, Wan, Sun, and Zhang]{gao2024honestllm}
Chujie Gao, Siyuan Wu, Yue Huang, Dongping Chen, Qihui Zhang, Zhengyan Fu, Yao Wan, Lichao Sun, and Xiangliang Zhang.
\newblock Honestllm: Toward an honest and helpful large language model.
\newblock \emph{Advances in Neural Information Processing Systems}, 37:\penalty0 7213--7255, 2024.

\end{thebibliography}

%%%%%%%%%%%%%%%%%%%%%%%%%%%%%%%%%%%%%%%%%%%%%%%%%%%%%%%%%%%%
%\clearpage

%\appendix

%\section{Appendix}

%\subsection{Role of Bias Term in Training and Inference}
%\label{sec:bias_dynamics}

%The bias term $b_i$ governs the activation of the transformation in both training and inference. At initialization, $\tilde{h}_\ell^{(i)} = h_\ell^{\text{ref}} + b_i$, producing a non-zero deviation $(\tilde{h}_\ell^{(i)} - h_\ell^{\text{ref}}) = b_i$, which activates the low-rank transformation in Equation (5). This allows gradients to propagate to $A_i$ and $B_i$ within the same forward computation (Lemma~2). During training, $b_i$, $A_i$, and $B_i$ are updated jointly. At inference, $b_i$ acts as a fixed offset, maintaining a non-zero deviation so the transformation remains active. The update is computed in a single forward pass without iteration or fixed-point refinement.

%%%%%%%%%%%%%%%%%%%%%%%%%%%%%%%%%%%%%%%%%%%%%%%%%%%%%%%%%%%%
\clearpage
\newpage

\end{document}